\documentclass[runningheads]{llncs}
\usepackage[T1]{fontenc}
\usepackage{graphicx,verbatim}
\usepackage{booktabs, multirow, xcolor}
\usepackage{array}
\usepackage{float}
\usepackage{microtype}
\usepackage{amsmath, amssymb, amsfonts}
\usepackage[export]{adjustbox}
\begin{document}
\title{PathGLS: Evaluating Pathology Vision-Language Models without Ground Truth through Multi-Dimensional Consistency}
\titlerunning{PathGLS: Reference-free Evaluation for Pathology VLMs}
\author{Minbing Chen \and Zhu Meng\textsuperscript{*} \and Fei Su\thanks{Corresponding authors.}}
\institute{Beijing University of Posts and Telecommunications \\
\email{\{cmb, bamboo, sufei\}@bupt.edu.cn}}
\authorrunning{Minbing Chen, Zhu Meng, and Fei Su}
\maketitle

\begin{abstract}
Vision-Language Models (VLMs) offer significant potential in computational pathology by enabling interpretable image analysis, automated reporting, and scalable decision support. However, their widespread clinical adoption remains limited due to the absence of reliable, automated evaluation metrics capable of identifying subtle failures such as hallucinations. To address this gap, we propose PathGLS, a novel reference-free evaluation framework that assesses pathology VLMs across three dimensions: \textit{Grounding} (fine-grained visual-text alignment), \textit{Logic} (entailment graph consistency using Natural Language Inference), and \textit{Stability} (output variance under adversarial visual-semantic perturbations). PathGLS supports both patch-level and whole-slide image (WSI)-level analysis, yielding a comprehensive trust score. Experiments on Quilt-1M, TCGA, REG2025, PathMMU and TCGA-Sarcoma datasets demonstrate the superiority of PathGLS. Specifically, on the Quilt-1M dataset, PathGLS reveals a steep sensitivity drop of 40.2\% for hallucinated reports compared to only 2.1\% for BERTScore. Moreover, validation against expert-defined clinical error hierarchies reveals that PathGLS achieves a strong Spearman's rank correlation of $\rho=0.71$ ($p < 0.0001$), significantly outperforming Large Language Model (LLM)-based approaches (Gemini 3.0 Pro: $\rho=0.39, p < 0.0001$). These results establish PathGLS as a robust reference-free metric. By directly quantifying hallucination rates and domain shift robustness, it serves as a reliable criterion for benchmarking VLMs on private clinical datasets and informing safe deployment. Code can be found at : \url{https://github.com/My13ad/PathGLS}

\end{abstract}
\section{Introduction}

\begin{figure}[!htb]
\centering
\includegraphics[width=\textwidth]{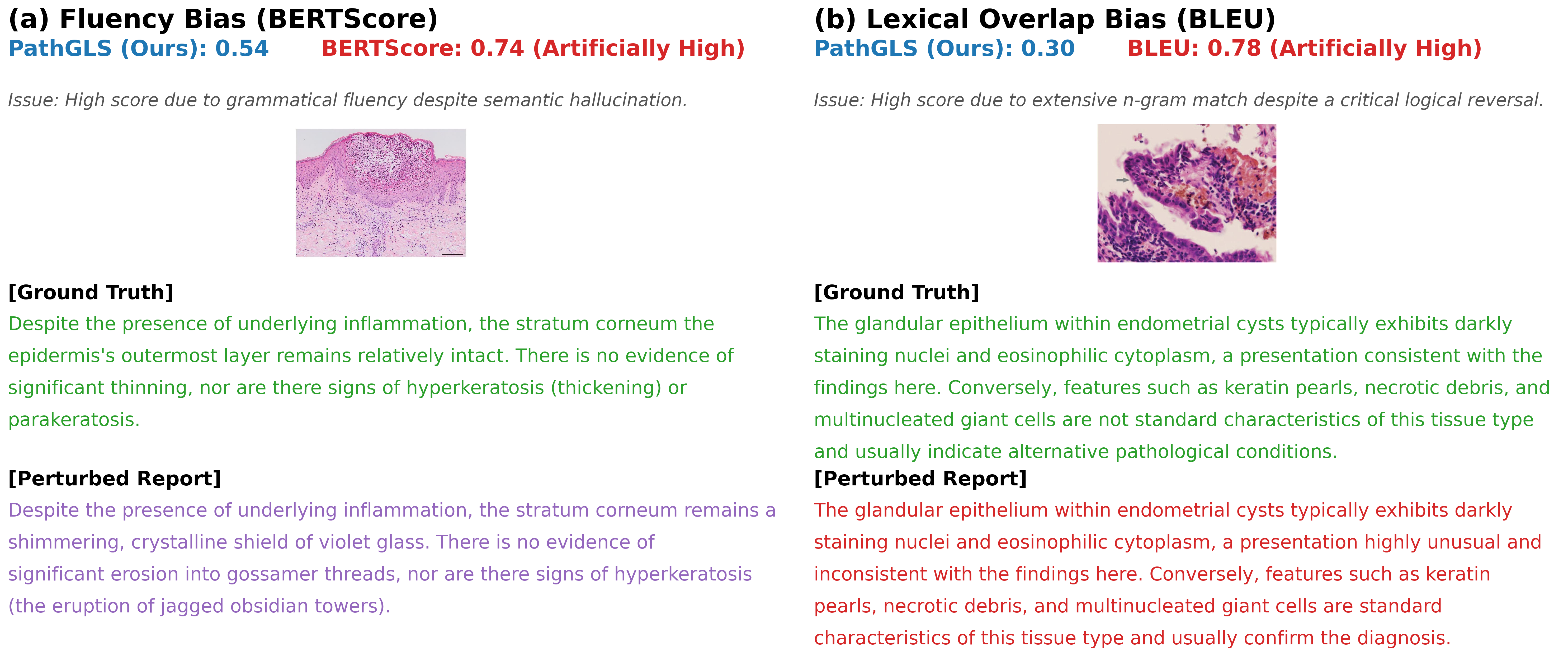} 
\caption{Vulnerabilities of traditional metrics on LLM-perturbed reports. \textbf{(a)} BERTScore exhibits \textit{fluency bias}, assigning artificially high scores to fluent but semantic hallucinations. \textbf{(b)} BLEU exhibits \textit{lexical overlap bias}, failing to penalize logical reversals. PathGLS effectively detects both errors.}
\label{fig:counter_intuitive}
\end{figure}

The shift towards Vision-Language Models (VLMs) in computational pathology offers generative reporting for clinical decision support. However, current VLMs frequently suffer from a dichotomy between \textit{fluency} and \textit{factuality}, generating grammatically perfect but semantically fabricated reports. Because perfect expert-annotated ground truths are rarely available for every whole-slide image (WSI), traditional reference-based metrics (e.g., BLEU~\cite{papineni2002bleu}, BERTScore~\cite{zhang2019bertscore}) remain ineffective. As illustrated in Fig.~\ref{fig:counter_intuitive}, conventional metrics blindly reward lexical overlap and stylistic fluency but fail to penalize logical reversals or semantic hallucinations.

To address this, we introduce PathGLS, a reference-free evaluation framework designed to quantify trust in pathology VLMs. Our core contributions are: (1) PathGLS, a multi-dimensional consistency evaluation protocol is proposed to quantify VLMs trustworthiness from three complementary perspectives: visual-textual grounding, logical consistency, and adversarial stability. (2) A dual adversarial attack strategy is introduced to systematically assess model robustness under clinical distribution shifts via stain perturbation and semantic injection. (3) PathGLS supports both patch-level and whole-slide image (WSI)-level evaluation, where WSI-level grounding is achieved through a high-resolution multiple instance learning (MIL) alignment mechanism that preserves diagnostic details. (4) Extensive experiments on multiple public and multi-center datasets (Quilt-1M\cite{ikezogwo2023quilt}, PathMMU\cite{sun2024pathmmu}, TCGA\cite{weinstein2013cancer}, REG2025, TCGA-Sarcoma\cite{weinstein2013cancer}) demonstrate that the proposed  PathGLS significantly outperforms existing metrics (e.g., BERTScore\cite{zhang2019bertscore}, BLEU\cite{papineni2002bleu}, RadGraph\cite{jain2021radgraph}, and LLM-as-a-judge) in detecting model hallucinations, exposing the fluency bias and logical reversal blind spots inherent in traditional metrics.

\section{Related Work}
Medical VLMs~\cite{li2023llavamed,ikezogwo2023quilt} enable generative reporting but suffer from fluent hallucinations. Evaluating these failures is challenging: traditional metrics~\cite{papineni2002bleu,zhang2019bertscore} exhibit fluency bias, while general hallucination benchmarks~\cite{li2023evaluating,rohrbach2018object} lack histopathological granularity. Furthermore, specialized medical metrics like RadGraph~\cite{jain2021radgraph} focus heavily on text-to-text extraction, ignoring the underlying image data. Consequently, existing text-centric methods fail to detect critical grounding errors where generated text visually contradicts the slide. PathGLS addresses this gap via a reference-free, multi-dimensional metric explicitly enforcing visual-text alignment and logical consistency.
\section{Methodology}

\subsection{Framework Overview}
\begin{figure}[!htb]
\centering
\includegraphics[width=\textwidth]{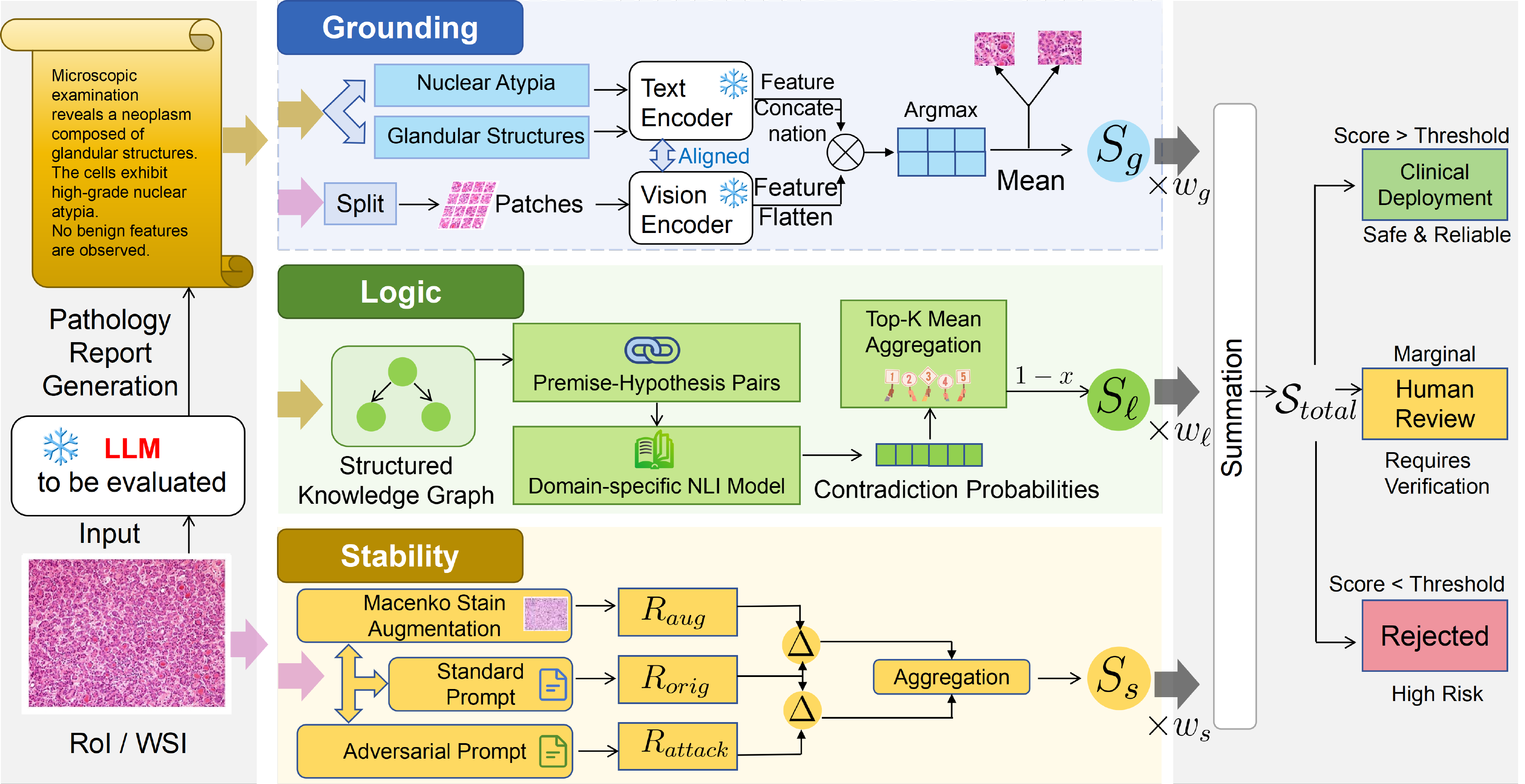} 
\caption{The overall architecture of PathGLS.}
\label{fig:architecture}
\end{figure}

To address the lack of ground truth in clinical settings, we propose PathGLS, a reference-free evaluation framework. As illustrated in Fig.~\ref{fig:architecture}, the target Vision-Language Model (Subject) generates a pathology report from an input ROI/WSI, which is then evaluated by an automated Judge System across three parallel dimensions: (1) \textbf{Grounding} ($S_g$) adopts a MIL strategy, utilizing a vision encoder and matrix multiplication to align fine-grained patch features with text embeddings, validating visual evidence via spatial argmax and mean pooling; (2) \textbf{Logic} ($S_\ell$) extracts premise-hypothesis pairs via a Structured Knowledge Graph and utilizes a domain-specific NLI model to compute contradiction probabilities, applying Top-K mean aggregation to penalize logical hallucinations; (3) \textbf{Stability} ($S_s$) quantifies robustness by computing semantic distances ($\Delta$) between the original report and those generated under visual (Macenko) and textual (adversarial) perturbations. Ultimately, these three metrics are fused into a comprehensive score via a weighted combination ($\mathcal{S}_{total} = S_g \times w_g + S_\ell \times w_\ell + S_s \times w_s$). This score serves as a clinical decision guardrail to guide the routing of the VLM outputs for deployment, human review, or rejection.

\subsection{Grounding Module: High-Resolution Multiple Instance Learning Alignment}
Standard vision-language metrics typically resize images to low resolutions, resulting in the loss of critical diagnostic features like nuclear atypia. Leveraging the MIL paradigm, we design a high-resolution alignment mechanism. 
The input ROI/WSI is tessellated into a bag of $N$ patches. A pathology-specific vision encoder extracts visual embeddings $v_i \in \mathbb{R}^D$ for each patch. Simultaneously, $M$ clinical entities are extracted from the generated report and encoded into text embeddings $t_j \in \mathbb{R}^D$. To establish cross-modal alignment without losing spatial granularity, we compute an $M \times N$ similarity matrix via matrix multiplication. The grounding score $S_g$ is derived by applying a spatial \textit{argmax} to identify the most relevant patch for each text entity, followed by a mean aggregation over all $M$ entities:

\begin{equation}
\label{eq:grounding}
S_g = \frac{1}{M} \sum_{j=1}^{M} \max_{1 \le i \le N} (v_i^\top t_j),
\end{equation}
which ensures that every clinical claim is objectively grounded by at least one specific visual region within the WSI bag.

\subsection{Logic Module: Graph-based Consistency Check}
This module evaluates the internal self-consistency of the generated report. A Natural Language Inference (NLI) approach combined with graph construction is employed.

First, the unstructured pathology report is parsed into a structured knowledge graph, where nodes represent medical entities and edges represent relations. To systematically verify reasoning chains, we extract premise-hypothesis pairs from this graph, typically pairing morphological descriptions (premise) with the final diagnosis (hypothesis). A domain-specific NLI model evaluates these pairs to output continuous contradiction probabilities. 

To prevent severe logical hallucinations from being diluted by a large number of consistent statements, we avoid global averaging and instead apply a top-$K$ mean aggregation mechanism. The final logic score $S_\ell$ is calculated by averaging the top $K$ most contradictory pairs, formulated as:

\begin{equation}
\label{eq:logic}
S_\ell = 1 - \frac{1}{K} \sum_{k=1}^{K} p_{(k)},
\end{equation}
where $p_{(k)}$ denotes the $k$-th highest contradiction probability among all evaluated pairs. This mechanism explicitly penalizes broken reasoning chains, ensuring that all diagnostic conclusions are logically entailed by the underlying morphological evidence.

\subsection{Stability Module: Adversarial Robustness}
To further test the model's reliability, an adversarial evaluation protocol is introduced, comprising two distinct attack vectors mapped in our framework. (1) Visual Perturbation (Macenko Stain Augmentation): Since pathology slides vary in staining, a robust model should maintain consistency. We apply a color deconvolution-based stain normalization technique to perturb the stain vectors, generating an augmented view. (2) Semantic Attack (Adversarial Prompt): To test diagnostic convictions, an adversarial prompt containing a false clinical history is injected to induce a cognitive bias.

The stability score $S_s$ is defined by the semantic consistency between the report generated from the original input ($R_{orig}$) and the reports from the perturbed inputs ($R_{aug}$ and $R_{attack}$). To mathematically ensure non-negativity, the semantic distances are constrained by absolute values before the mean aggregation: 
\begin{equation}
\label{eq:stability}
    S_s = 1 - \frac{1}{2} \left[ |\Delta(R_{orig}, R_{aug})| + |\Delta(R_{orig}, R_{attack})| \right],
\end{equation}
where $|\Delta(\cdot, \cdot)|$ represents the absolute semantic distance normalized to the range $[0, 1]$. A high stability score indicates strong robustness against both domain shifts and cognitive bias induction.

\section{Experiments}
\subsection{Datasets and Implementation Details}
\textbf{Datasets:} We utilize five datasets. Patch-level benchmarks employ Quilt-1M~\cite{ikezogwo2023quilt} and PathMMU~\cite{sun2024pathmmu}. WSI-level benchmarks use TCGA~\cite{weinstein2013cancer} and the REG2025 dataset (20,500 WSI-report pairs across seven organs from six centers) to rigorously evaluate generalization. To assess robustness against domain shifts, we curated an Out-of-Distribution (OOD) proxy subset from TCGA-Sarcoma~\cite{weinstein2013cancer}, leveraging its high morphological variance. Finally, for sensitivity analysis, we curated a Perturbed Dataset by modifying ground truth captions to create Control, Visual Hallucination, and Logic Error groups.

\textbf{Implementation Details:} WSIs are processed via a sliding window to extract 512$\times$512 patches at 20$\times$ magnification. The Grounding, Logic, and Stability modules utilize HighRes-PLIP~\cite{huang2023visual} (stride 224), DeBERTa-v3-base~\cite{he2021debertav3}, and Macenko normalization~\cite{macenko2009method} as their respective backbones. Trust score weights ($w_g=0.4, w_l=0.3, w_s=0.3$) are determined via grid search on a hold-out validation set to prioritize visual accuracy. All experiments are conducted on a single NVIDIA RTX 4090 GPU.

\subsection{Metric Validation: Reliability and Sensitivity}
\begin{figure}[!htb]
\centering
\begin{minipage}{0.48\textwidth}
  \centering
  \includegraphics[width=\linewidth, trim=0 0 0 1.6cm, clip]{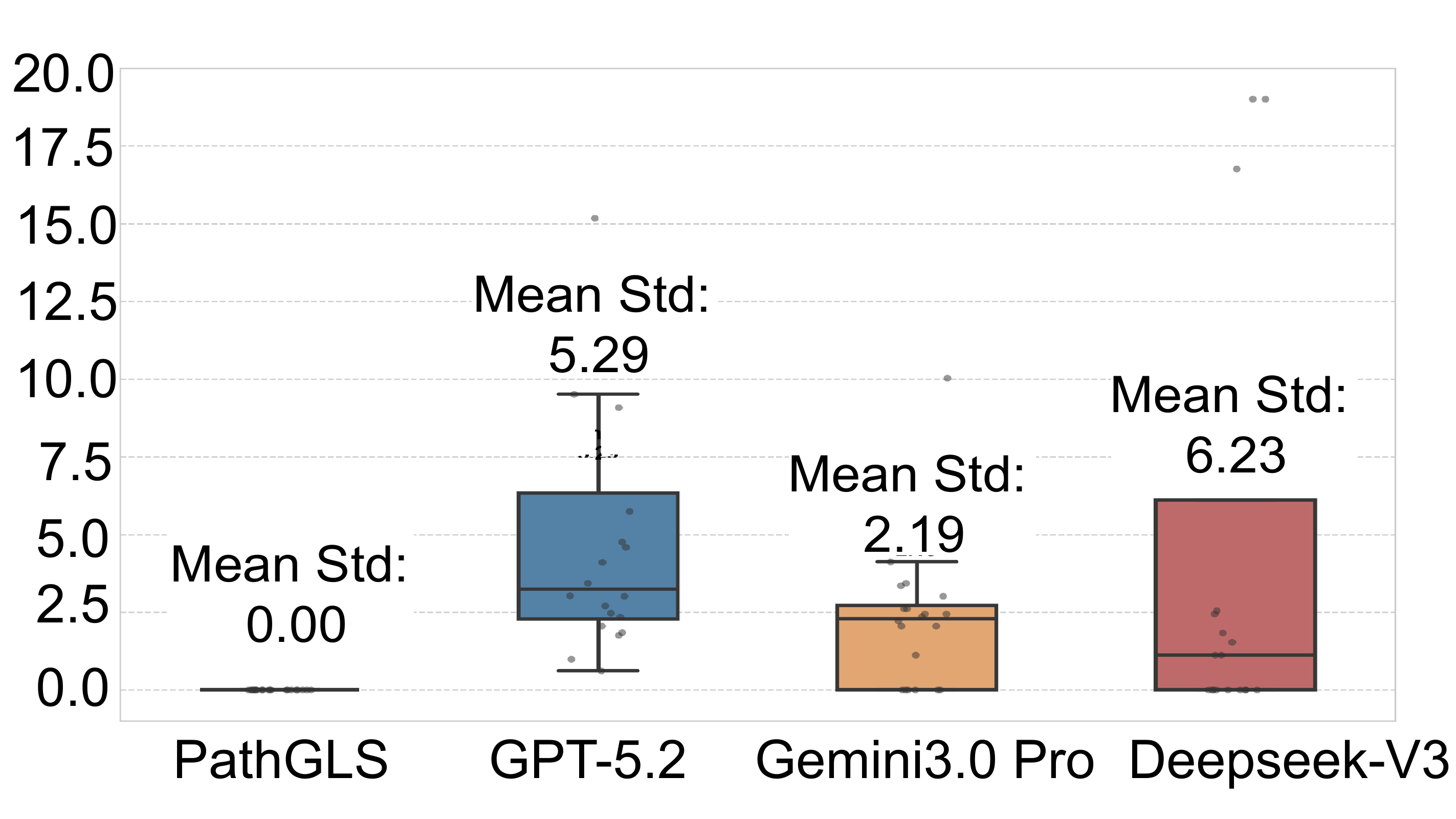}
  \centerline{\small (a) Robustness Analysis}
\end{minipage}\hfill
\begin{minipage}{0.48\textwidth}
  \centering
  \includegraphics[width=\linewidth, trim=0 0 0 2cm, clip]{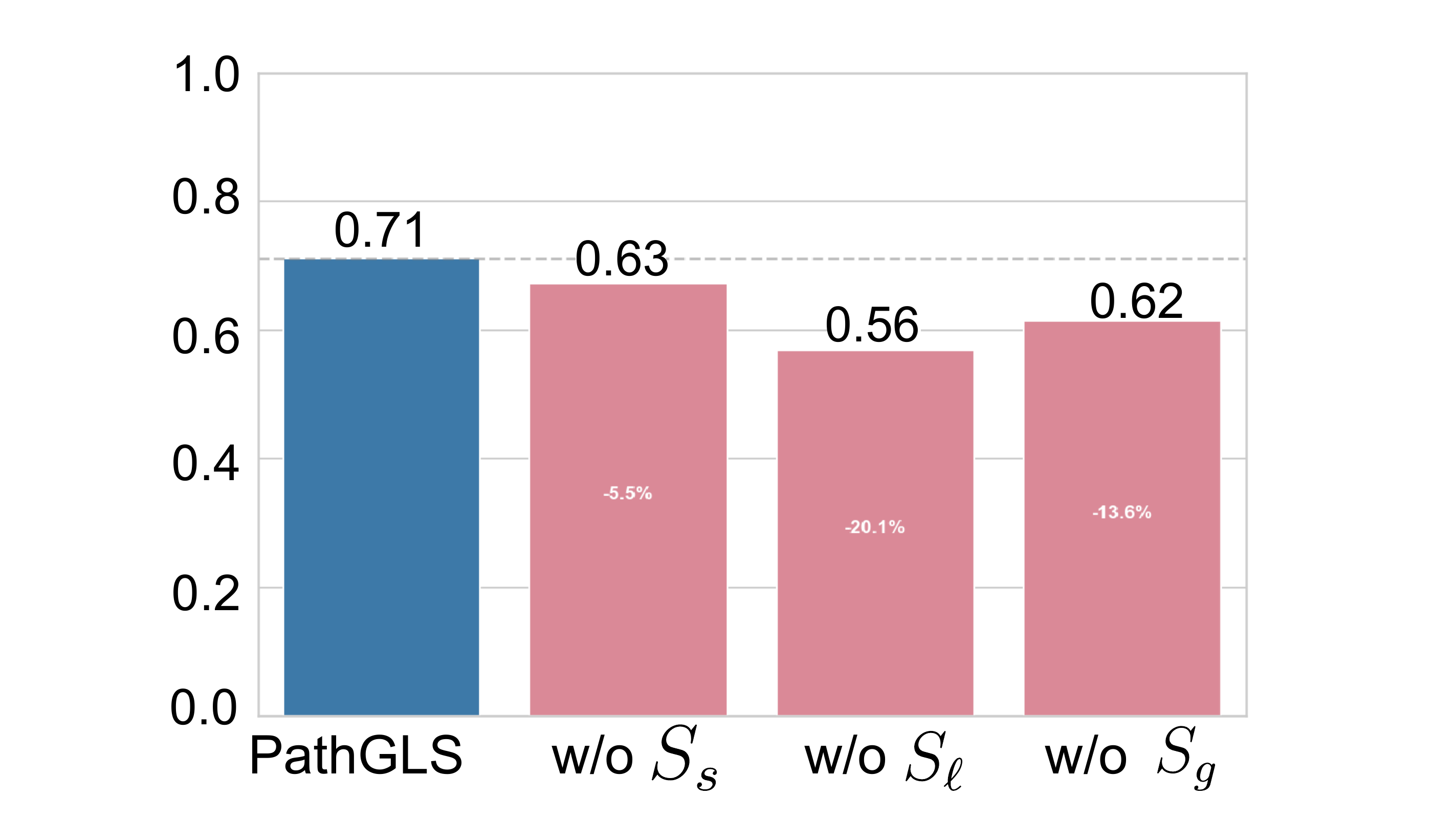}
  \centerline{\small (b) Ablation Study}
\end{minipage}
\caption{Metric Validation. (a) LLM judges show high variance, whereas PathGLS demonstrates perfect stability. (b) Logic contributes most significantly to the final score.}
\label{fig:validation_combined}
\end{figure}

As shown in Table~\ref{tab:sensitivity}, traditional metrics like BERTScore exhibit a severe fluency bias, maintaining high scores (0.92 $\rightarrow$ 0.90) for hallucinated outputs. Conversely, PathGLS demonstrates a sharp sensitivity gradient: $S_g$ drops by 40.2\% for Visual Hallucinations, and $S_\ell$ drops by 26.4\% for Logic Errors. (Note: Stability $S_s$ evaluates dynamic generation variance under perturbation, and is thus structurally excluded from this static text experiment). Furthermore, while LLM-as-a-judge yields high variance rendering it unreliable (Fig.~\ref{fig:validation_combined}a), PathGLS provides deterministic stability (Std=0.00). An end-to-end ablation study (Fig.~\ref{fig:validation_combined}b) confirms the essential contribution of all modules to human alignment: removing Logic, Grounding, or Stability decreases the metric's Spearman correlation with pre-defined error hierarchy by 20.1\%, 13.6\%, and 5.5\%, respectively.
\begin{table}[!htb]
\centering
\footnotesize
\caption{Comparison of sensitivity to hallucinations on Quilt-1M.}
\label{tab:sensitivity}
\begin{tabular}{l c | c | c | c c | c c}
\toprule
\multicolumn{2}{l|}{\multirow{2}{*}{\textbf{Metric}}} & \multirow{2}{*}{\textbf{Focus}} & \multirow{2}{*}{\textbf{Control}} & \multicolumn{2}{c|}{\textbf{Visual Hallucination}} & \multicolumn{2}{c}{\textbf{Logic Error}} \\
\cmidrule(lr){5-6} \cmidrule(lr){7-8}
\multicolumn{2}{l|}{} & & & \textbf{Score} & \textbf{$\Delta\%$} & \textbf{Score} & \textbf{$\Delta\%$} \\
\midrule
\multicolumn{2}{l|}{BLEU-4~\cite{papineni2002bleu}} & Lexical & 0.16 & 0.12 & 25.0\% & 0.13 & 18.8\% \\
\multicolumn{2}{l|}{RadGraph~\cite{jain2021radgraph}} & Entity & 0.31 & 0.19 & 38.7\% & 0.25 & 19.4\% \\
\multicolumn{2}{l|}{BERTScore~\cite{zhang2019bertscore}} & Semantic & 0.92 & 0.90 & 2.2\% & 0.91 & 1.1\% \\
\midrule
\multirow{2}{*}{PathGLS} & $S_g$ & Visual-Text & 0.77 & 0.46 & \textbf{40.3\%} & 0.73 & 5.2\% \\
& $S_l$ & Consistency & 0.91 & 0.82 & 9.9\% & 0.67 & \textbf{26.4\%} \\
\bottomrule
\end{tabular}
\end{table}

\subsection{Benchmarking and Scale Sensitivity}
We benchmark models across diverse datasets (Table~\ref{tab:benchmark}). Quilt-LLaVA consistently outperforms LLaVA-Med in overall PathGLS. For instance, on Quilt-1M Grounding, Quilt-LLaVA achieves 0.42 versus LLaVA-Med's 0.38. Transitioning to WSI-level increases overall PathGLS for both models; however, logic consistency trends diverge. LLaVA-Med maintains stable Logic (0.97 on TCGA), whereas Quilt-LLaVA drops to 0.78. This reveals that while domain-specific pre-training improves aggregate performance, maintaining logical coherence across disjointed WSI regions remains challenging.

\begin{table}[!htb]
\centering
\footnotesize
\caption{Benchmarking results.}
\label{tab:benchmark}
\begin{tabular}{l|l|c|ccc|c}
\toprule
\textbf{Dataset} & \textbf{Model} & \textbf{Scale} & \textbf{Grounding} & \textbf{Logic} & \textbf{Stability} & \textbf{PathGLS} \\ 
\midrule
\multirow{3}{*}{Quilt-1M} & LLaVA-Med & Patch & 0.38 & 0.96 & 0.30 & 0.53 \\
 & Quilt-LLaVA & Patch & 0.42 & 0.92 & 0.52 & 0.60 \\ 
 & MedGemma-4B-it\cite{sellergren2025medgemma} & Patch & 0.49 & 0.91 & 0.40 & 0.59 \\
\midrule
\multirow{3}{*}{PathMMU} & LLaVA-Med & Patch & 0.46 & 0.96 & 0.62 & 0.65 \\
 & Quilt-LLaVA & Patch & 0.46 & 0.97 & 0.72 & 0.69 \\ 
 & MedGemma-4B-it & Patch & 0.71 & 0.97 & 0.75 & 0.80 \\
\midrule
\multirow{3}{*}{TCGA} & LLaVA-Med & WSI & 0.73 & 0.97 & 0.75 & 0.80 \\
 & Quilt-LLaVA & WSI & 0.96 & 0.78 & 0.73 & 0.83 \\ 
 & MedGemma-4B-it & WSI & 0.50 & 0.91 & 0.32 & 0.56 \\
\midrule
\multirow{3}{*}{REG2025} & LLaVA-Med & WSI & 0.72 & 0.96 & 0.72 & 0.79 \\
 & Quilt-LLaVA & WSI & 0.72 & 0.97 & 0.74 & 0.80 \\ 
 & MedGemma-4B-it & WSI & 0.64 & 0.98 & 0.62 & 0.74 \\
\bottomrule
\end{tabular}
\end{table}
\subsection{Clinical Deployment: Domain Gap on Unseen Cohorts}

Validating models on unseen private cohorts (e.g., REG2025) and rare subtypes (e.g., TCGA-Sarcoma) is critical for clinical safety. Traditional metrics fail in these scenarios by assigning high scores to fluent hallucinations, rendering them unsafe for automated benchmarking.
\begin{table}[htbp]
\centering
\caption{Domain Gap Analysis: Comparison of PathGLS scores between in-domain and out-of-domain datasets.}
\label{tab:domain_gap}
\begin{tabular}{lccc}
\hline
\textbf{Model} & \textbf{Public Dataset} & \textbf{Private Dataset} & \textbf{Drop ($\Delta$)} \\ \hline
LLaVA          & 0.801                  & 0.737                   & 0.064                    \\
Quilt-LLaVA    & 0.845                  & 0.836                   & 0.009                    \\ \hline
\end{tabular}
\end{table}

\noindent \textbf{PathGLS as a Clinical Gatekeeper.} PathGLS significantly outperforms traditional metrics in identifying reliable models under domain shifts (Table \ref{tab:domain_gap}). While BERTScore remains deceptively high, PathGLS accurately penalizes general-domain models (LLaVA) failing to generalize, exposing a significant grounding drop on unseen morphologies (PathGLS decreases by 0.064). Conversely, it validates the robustness of pathology-specific models (Quilt-LLaVA, $\Delta = 0.009$). Thus, PathGLS serves as a rigorous, reference-free criterion for VLM selection in proprietary clinical deployments.

\begin{table}[t] 
    \centering
    \fontsize{8pt}{9pt}\selectfont 
    \setlength{\tabcolsep}{1pt}
    
    \caption{Qualitative evaluation. \textcolor{red}{\textbf{Red}}: Hallucinations/logic errors (penalized by PathGLS, missed by BERTScore). \textcolor{green!60!black}{\textbf{Green}}: Accurate reasoning.}
    \label{tab:qualitative_results}
    
    \begin{tabular}{ 
        >{\centering\arraybackslash}m{0.11\textwidth} |
        >{\raggedright\arraybackslash}m{0.28\textwidth} | 
        >{\raggedright\arraybackslash}m{0.28\textwidth} | 
        >{\raggedright\arraybackslash}m{0.28\textwidth} 
    }
        \toprule
        \textbf{Dataset} & \textbf{Ground Truth} & \textbf{Quilt-LLaVA} & \textbf{LLaVA-Med} \\
        \midrule
        
        % ==================== PATHMMU ====================
        \textbf{Path\-MMU}\newline
        \includegraphics[width=0.8\linewidth, keepaspectratio]{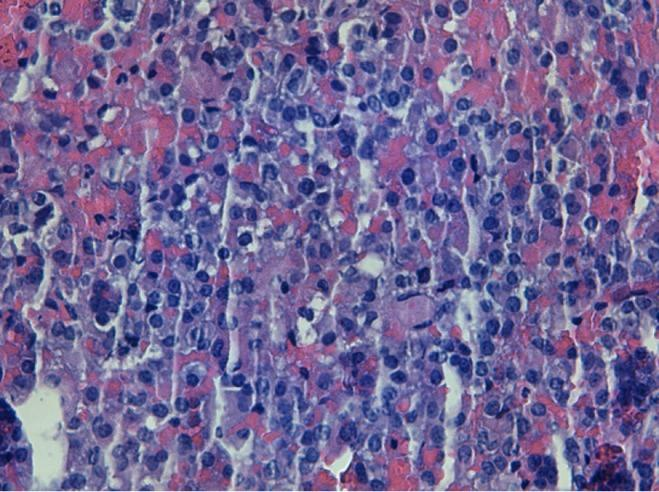} 
        & 
        \textbf{Diag:} Ovoid/polygonal cells. \newline
        \textbf{Features:} Monotonous, no overt atypia/mitoses.
        & 
        \textbf{PathGLS: 0.56} $|$ \textbf{BERT: 0.81} \newline
        \textcolor{red}{\textbf{Hepatocytes}} in cords. \textcolor{red}{\textbf{Kupffer cells}}. \newline
        $\rightarrow$ \textcolor{red}{\textbf{Normal liver histology}}.
        & 
        \textbf{PathGLS: 0.36} $|$ \textbf{BERT: 0.85} \newline
        \textcolor{red}{\textbf{Plasma cells}}, \textcolor{red}{\textbf{prominent nucleoli}}. \newline
        $\rightarrow$ \textcolor{red}{\textbf{Castleman disease}}. \\
        \midrule
        
        % ==================== QUILT1M ====================
        \textbf{Quilt-1M}\newline (Lymph Node)\newline
        \includegraphics[width=0.8\linewidth, keepaspectratio]{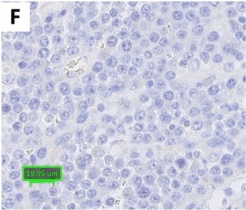} 
        & 
        \textbf{Diag:} Atypical lymphoid infiltrates. \newline
        \textbf{IHC:} CD79a+, MUM-1+, Ki67 $>$90\%.
        & 
        \textbf{PathGLS: 0.38} $|$ \textbf{BERT: 0.80} \newline
        Architecture \textcolor{red}{\textbf{well-preserved}}. \newline
        $\rightarrow$ \textcolor{red}{\textbf{Healthy, no malignancy}}.
        & 
        \textbf{PathGLS: 0.88} $|$ \textbf{BERT: 0.83} \newline
        \textcolor{green!60!black}{\textbf{Small/large lymphocytes}}. Diffuse. \newline
        $\rightarrow$ \textcolor{green!60!black}{\textbf{Diffuse large B-cell lymphoma}}. \\
        \bottomrule
    \end{tabular}
\end{table}

\textbf{Interpretable Evidence for Trust.} Beyond scalar scores, PathGLS provides granular interpretability. By decomposing performance into Grounding, Logic, and Stability, it offers specific evidence of model failures. As shown in Table~\ref{tab:qualitative_results}, PathGLS explicitly captures visual-textual disconnects in PathMMU example that BERTScore misses, establishing it as an interpretable framework for substantiating clinical decision support.

\section{Conclusion}
The deployment of VLMs in computational pathology faces a critical Trust Paradox, where high textual fluency frequently masks severe, clinically dangerous hallucinations. To address the failure of traditional metrics in penalizing these hidden errors, we propose PathGLS, a reference-free evaluation framework tailored specifically for pathology VLMs. By enforcing multi-dimensional consistency across visual grounding, logical reasoning, and diagnostic stability, PathGLS provides a holistic and robust quantification of clinical trustworthiness. It serves as a reliable criterion for benchmarking and selecting generalizable VLMs prior to real-world clinical deployment.
\section*{Acknowledgments}
This work is supported by Chinese National Natural Science Foundation (62401069)
\bibliographystyle{splncs04}
\bibliography{references}
\end{document}